\documentclass[twoside,11pt]{article}

\usepackage{blindtext}

\usepackage{jmlr2e}
\usepackage{hyperref}
\usepackage{booktabs}
\usepackage{xcolor}
\usepackage{lastpage}
\usepackage{algorithm}
\usepackage[noend]{algpseudocode}
\usepackage{amsmath, amssymb}
\usepackage{pgfplots}
\usepackage{makecell}

\pgfplotsset{compat=1.17}

\firstpageno{1}

\begin{document}

\title{An Adaptive Volatility-based Learning Rate Scheduler}

\author{\name Kieran Chai Kai Ren \email h2110069@nushigh.edu.sg \\
       \addr NUS High School of Mathematics and Science\\
       20 Clementi Ave 1, Singapore 129957
}
\maketitle

\begin{abstract}%   <- trailing '%' for backward compatibility of .sty file
Effective learning rate (LR) scheduling is crucial for training deep neural networks. However, popular pre-defined and adaptive schedulers can still lead to suboptimal generalization. This paper introduces VolSched, a novel adaptive LR scheduler inspired by the concept of volatility in stochastic processes like Geometric Brownian Motion to dynamically adjust the learning rate. By calculating the ratio between long-term and short-term accuracy volatility, VolSched increases the LR to escape plateaus and decreases it to stabilize training, allowing the model to explore the loss landscape more effectively. We evaluate VolSched on the CIFAR-100 dataset against a strong baseline using a standard augmentation pipeline. When paired with ResNet-18 and ResNet-34, our scheduler delivers consistent performance gains, improving top-1 accuracy by 1.4 and 1.3 percentage points respectively. Analysis of the loss curves reveals that VolSched promotes a longer exploration phase. A quantitative analysis of the Hessian shows that VolSched finds a final solution that is 38\% flatter than the next-best baseline, allowing the model to obtain wider minima and hence better generalization performance.
\end{abstract}

\section{Introduction}

During neural network training, learning rates often have to be changed over time as the model learns to fit the training data. The prior literature includes static, non-adaptive methods to schedule LRs, such as Cosine Annealing  \cite{cosineanneal}, along with simple adaptive methods such as ReduceLROnPlateau \cite{reducelr}. 

While pre-defined schedules like Cosine Annealing are effective, they are blind to the model's real-time progress. If the initial LR is too low, they will still decay it, potentially getting stuck. Adaptive methods like ReduceLROnPlateau can react, but only by decreasing the LR. VolSched addresses this gap by providing a mechanism to dynamically increase or decrease the LR in response to changes in training progress, adaptively raising the learning rate when training stalls and lowering it otherwise.

The proposed volatility-based scheduler provides several desirable properties. It is computationally lightweight, and can be simply integrated into standard training pipelines. We demonstrate its efficacy through an evaluation on CIFAR-100. Compared to strong baselines including ReduceLROnPlateau and Cosine Annealing, using modern data augmentation, our scheduler boosts the performance of ResNet-18 and ResNet-34 \cite{resnet} by 1.4 and 1.3 absolute percentage points in top-1 accuracy, respectively.

\section{Background}
\subsection{Geometric Brownian motion (GBM)}
A GBM is a stochastic process often used in the modeling of financial markets, in which the logarithm of the random variable is modeled as a Brownian motion with drift. This can be used to model the accuracy of a machine learning model, as it has an overall increasing trend with stochastic variations. A GBM can be defined as the following stochastic differential equation:
\[ dS = \mu Sdt + \sigma SdW \]
where $S$ is the model accuracy, $W$ is a Brownian motion, $\mu$ is the drift, and $\sigma$ is the percentage volatility. 
In the context of our GBM analogy, existing methods can be seen as reacting to the drift term ($\mu$). For instance, \cite{reducelr} lowers the learning rate upon $\mu$ reaching 0 for an extended period of time.

\section{VolSched}
\subsection{Volatility ratio calculation}
Our proposed method makes use of the $\sigma$ term in the GBM equation. We treat the per-batch training accuracy analogously to a GBM, then use the logarithm of the percentage volatility to scale the LR accordingly. We chose training accuracy over loss due to its monotonic trends and reduced sensitivity to local noise in early training stages. While not a true GBM, its variance can heuristically capture training progress. We use the following algorithm:

\begin{algorithm}
\caption{Calculate volatility ratio}
\label{alg:transformed_ratio}
\begin{algorithmic}[1]
\Require
    $L = \{l_1, l_2, \dots, l_T\}$, a sequence of historical per-batch train accuracy.
    \Statex \hspace{\algorithmicindent} $N \in \mathbb{Z}^+$, the volatility window size.
    \Statex \hspace{\algorithmicindent} $w \ge 0$, a non-negative weight constant.
    \Statex \hspace{\algorithmicindent} $\varepsilon > 0$, a small constant for numerical stability (e.g., $10^{-8}$).
\Ensure
    $M \in \mathbb{R}$, the final transformed multiplier.
\Statex
\Procedure{CalculateVolRatio}{$L, N, w, \varepsilon$}
    \State Let $T$ be the number of elements in $L$.
    \State Let $L'=\{l'_1, \dots, l'_T\}$ where $l'_i \gets \max(l_i, \varepsilon)$ for $i=1, \dots, T$.
    \State Let $R = \{r_1, \dots, r_{T-1}\}$ where $r_i \gets \ln(l'_{i+1}/l'_i)$. \Comment{Calculate log-returns on clipped values}
    \State Let $R_N = \{r_{T-N}, \dots, r_{T-1}\}$ be the last $N$ elements of $R$.
    \Statex
    \State $\sigma_{\text{all}} \gets \text{StDev}(R)$ \Comment{Standard deviation of all log-returns}
    \State $\sigma_N \gets \text{StDev}(R_N)$ \Comment{Standard deviation of recent log-returns}
    \Statex
    \State $\rho \gets \sigma_{\text{all}} / \sigma_N$ \Comment{Compute the volatility ratio}
    \State $\delta \gets \rho - 1$ \Comment{Calculate the signed deviation from 1}
    \State $\Delta \gets \text{sgn}(\delta) \cdot \ln(1 + w \cdot |\delta|)$ \Comment{Apply the signed log-transformation}
    \State $M \gets 1 + \Delta$ \Comment{Calculate the final multiplier}
    \Statex
    \State \Return $M$
\EndProcedure
\end{algorithmic}
\end{algorithm}
\newpage

When the model is in a plateau, the accuracy value stagnates. This results in low short-term standard deviation $\sigma_N$. This, in turn, yields a high volatility ratio $\rho > 1$, triggering an increase in the learning rate to escape the plateau. We also apply a signed logarithmic transformation to dampen the effect of extreme volatility ratios, ensuring a more stable update to the learning rate. 

We introduce user defined constants $N$ and $w$, which represent update frequency and strength of response respectively. The scheduler performs a learning rate update every $N$ steps. Generally, lower $N$ can be used for larger batch size, while lower $w$ should be used for more difficult optimization problems. Deeper models like ResNet-34 may benefit from a more conservative volatility response to maintain training stability. We have experimentally determined $w = 0.03$ and $w = 0.05$ to work well for ResNet-34 and ResNet-18 \cite{resnet} respectively (see \hyperref[sec:hsa]{Appendix C} for a hyperparameter sensitivity analysis).

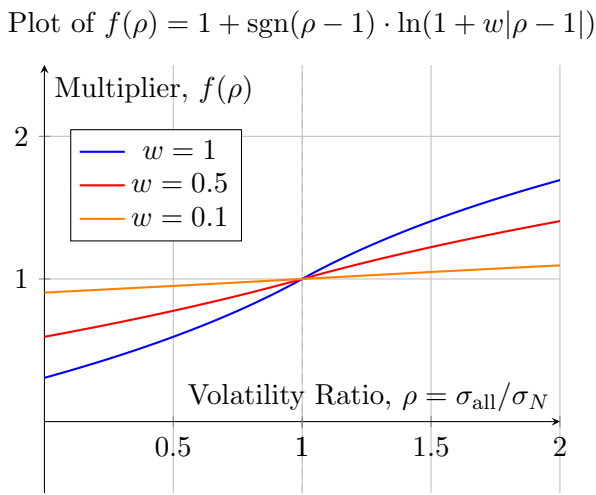
\begin{figure}[h!]
\centering
\begin{tikzpicture}
    \begin{axis}[
        title={Plot of $f(\rho) = 1 + \text{sgn}(\rho-1) \cdot \ln(1+w|\rho-1|)$},
        xlabel={Volatility Ratio, $\rho = \sigma_{\text{all}} / \sigma_N$},
        ylabel={Multiplier, $f(\rho)$},
        xmin=0, xmax=2,        % Set x-axis range
        ymin=-0.5, ymax=2.5,   % Set y-axis range
        grid=major,            % Add a major grid
        legend style={
            at={(0.05, 0.85)}, % Position at 95% right, 5% up
            anchor=north west
        },
        % Add dashed lines to highlight the axes and the critical point at rho=1
        extra x ticks={1},
        extra x tick style={grid=major, dashed},
        axis y line=middle,
        axis x line=middle,
    ]

    % The function to plot. pgfplots uses `x` as the variable.
    % The `sign()` function is available in the math parser.
    \addplot[
        blue, 
        smooth, 
        thick, 
        domain=0:4, 
        samples=200
    ] {1 + sign(x-1) * ln(1 + 1 * abs(x-1))};
    \addlegendentry{$w=1$}

    \addplot[
        red, 
        smooth, 
        thick, 
        domain=0:4, 
        samples=200
    ] {1 + sign(x-1) * ln(1 + 0.5 * abs(x-1))};
    \addlegendentry{$w=0.5$}

    \addplot[
        orange, 
        smooth, 
        thick, 
        domain=0:4, 
        samples=200
    ] {1 + sign(x-1) * ln(1 + 0.1 * abs(x-1))};
    \addlegendentry{$w=0.1$}

    \end{axis}
\end{tikzpicture}
\caption{The plot shows how the multiplier changes with the volatility ratio $\rho$ for different weight constants $w$. A larger $w$ leads to a more sensitive response.}
\label{fig:transformed_deviation}
\end{figure}

\subsection{Cosine Annealing}

While the volatility multiplier $M$ provides local, adaptive adjustments, a global decay schedule is still critical for eventual convergence. A naive approach would be to modulate a standard, stateless scheduler (like Cosine Annealing) with $M$ at each step. However, this would render the effects of $M$ transient, as a boost in learning rate at one step would be forgotten at the next, as the schedule would be recalculated from the initial learning rate.

To enable the effects of our volatility adaptation to persist and compound over time, we design VolSched as a multiplicative scheduler, where the new learning rate is a function of the previous learning rate:

\[\eta_{new} = \eta_{old}\cdot M \cdot \alpha\]

In this formulation, $M$ directly compounds its effect on the learning rate. If the model remains in a plateau ($M > 1$ repeatedly), or an unstable phase ($M < 1$) for multiple update steps, the learning rate will adjust exponentially, allowing for more aggressive exploration or training stabilization.

The role of the multiplier $\alpha$ is to simultaneously impose a global Cosine Annealing decay onto this multiplicative process. It acts as a correction factor that applies the amount of decay that should have occurred according to a standard cosine schedule over the last $N$ steps.

We define a base cosine decay function $g(t) = 1 + \cos\left(\frac{t}{T_{\max}}\pi\right)$. To ensure our scheduler tracks this base decay when $M=1$, the correction factor $\alpha$ must be the ratio of the base schedule's value at the current step to its value at the previous update step:

\[\alpha = \frac{g(T_{cur})}{g(T_{cur} - N)} = \frac{1 + \cos\left(\frac{T_{cur}}{T_{\max}}\pi\right)}{1 + \cos\left(\frac{T_{cur} - N}{T_{\max}}\pi\right)}\]

where $T_{cur}$ is the current step, $T_{\max}$ is the total number of steps, and $N$ is the volatility window size. This design decouples the adaptive, history-dependent modulation ($M$) from the global, time-dependent decay ($\alpha$), creating an effective schedule that can both explore aggressively and converge reliably.

\section{Experiments}
\subsection{ResNet on CIFAR-100}
We conducted experiments to evaluate various LR schedulers with ResNet-18 and ResNet-34 on CIFAR-100. We compare VolSched to Cosine Annealing, ExponentialLR, and ReduceLROnPlateau, with LR warmups. The configuration was trained multiple times with different random seeds to ensure robustness of the results. We apply RandomCrop, RandomHorizontalFlip, and AutoAugment \cite{autoaug} using the learned CIFAR-10 policy (see \hyperref[sec:exp]{Appendix A} for the full experimental setup). Hyperparameters were taken from the training recipes used at \cite{recipes}.

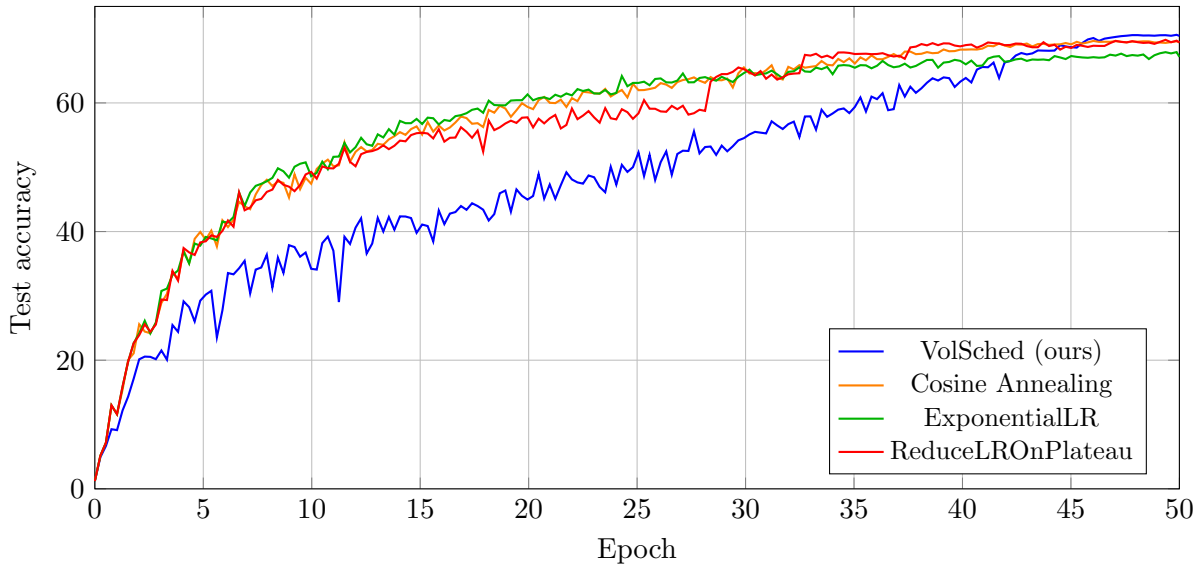
\begin{figure}[h!]
\centering
\begin{tikzpicture}
    \begin{axis}[
        % --- General Axis Settings ---
        width=16cm,
        height=8cm,
        grid=major,
        %
        % --- Axis Labels and Title ---
        xlabel={Epoch},
        ylabel={Test accuracy},
        %
        % --- Ticks and Limits ---
        xmin=0, xmax=50,
        ymin=0, ymax=75,
        %
        % --- Legend ---
        legend pos=south east,
        legend style={font=\small},
    ]

    % --- Plotting the Data from the CSV file ---

    % 1. Cosine Annealing (Blue)
    \addplot[no marks, blue, thick] 
    table [x=step, y=my_scheduler, col sep=comma] {my_data.csv};
    \addlegendentry{VolSched (ours)}

    % 2. My Scheduler (Orange)
    \addplot[no marks, orange, thick] 
    table [x=step, y=cosine, col sep=comma] {my_data.csv};
    \addlegendentry{Cosine Annealing}

    % 3. Exponential Scheduler (Green)
    \addplot[no marks, green!70!black, thick] 
    table [x=step, y=exponential, col sep=comma] {my_data.csv};
    \addlegendentry{ExponentialLR}

    % 4. Plateau Scheduler (Red)
    \addplot[no marks, red, thick] 
    table [x=step, y=plateau, col sep=comma] {my_data.csv};
    \addlegendentry{ReduceLROnPlateau}

    \end{axis}
\end{tikzpicture}
\caption{The plot shows the test accuracy on CIFAR-100 with various LR schedulers}
\label{fig:accs}
\end{figure}

\begin{table}[htbp]
\centering
\begin{tabular}{@{}llcccc@{}}
\toprule
\thead{Model} & \thead{Epochs} & \thead{VolSched} & \thead{Cosine\\Annealing} & \thead{ExponentialLR} & \thead{ReduceLROnPlateau} \\
\midrule
ResNet-18 & 20 & \textcolor{magenta}{\textbf{64.38$\pm$0.5}} & 63.33$\pm$0.1 & 60.06$\pm$1.1 & 57.20$\pm$0.6 \\
ResNet-18 & 50 & \textcolor{magenta}{\textbf{70.89$\pm$0.3}} & 69.31$\pm$0.2 & 67.35$\pm$0.5 & 69.47$\pm$0.7 \\
ResNet-34 & 50 & \textcolor{magenta}{\textbf{70.97$\pm$0.6}} & 69.30$\pm$0.3 & 67.16$\pm$0.3 & 69.66$\pm$0.3 \\
\bottomrule
\end{tabular}
\caption{Top-1 accuracy (\%) of ResNet-18 and ResNet-34 models on CIFAR-100. \textcolor{magenta}{\textbf{Best models}} are highlighted.}
\label{tab:results}
\end{table}

Overall, VolSched delivers gains over the baseline models for varying training durations. On CIFAR-100 with a ResNet-18 backbone, accuracy improves from 69.47\% to 70.89\%. Performing the paired t-test results in a statistically significant p value of $p = 0.0071$. 

ResNet-34 shows a similar boost, with accuracy improving from 69.66\% to 70.97\%. Furthermore, an improvement is also observed when training for fewer epochs, suggesting VolSched is not just a strategy that can be used exclusively for extended training runs.

\subsection{Swin Transformer}
To test the architectural generality of VolSched, we extended our evaluation to the attention-based Swin Transformer \cite{swin}. As is common for training transformer models, we used the AdamW \cite{adamw} optimizer instead of SGD for these experiments. For this, we conducted four paired-seed experiments, where both the baseline and VolSched models started from identical initial weights for each run. 
\begin{table}[htbp]
\centering
\begin{tabular}{@{}lcccc@{}}
\toprule
 & \thead{VolSched} & \thead{Cosine\\Annealing} & \thead{ExponentialLR} & \thead{ReduceLROnPlateau} \\
\midrule
Accuracy & \textcolor{magenta}{\textbf{57.22$\pm$0.4}} & 56.70$\pm$0.4 & 54.65$\pm$0.3 & 55.37$\pm$0.5 \\
\bottomrule
\end{tabular}
\caption{Top-1 accuracy (\%) of Swin Transformer models on CIFAR-100. \textcolor{magenta}{\textbf{Best model}} is highlighted.}
\label{tab:results}
\end{table}

While the average improvement was 0.52\%, VolSched consistently outperformed the baseline in every single run. Furthermore, performing the paired t-test also results in a statistically significant p value of $p = 0.024$. This demonstrates a robust and reliable advantage, suggesting that the benefits of volatility-based scheduling are a systematic improvement, even on architectures fundamentally different from CNNs.

A primary reason for the more modest gains observed with the Swin Transformer is likely due to architectural differences. Research suggests that transformers, due to their weaker inductive biases, often converge to wider, flatter minima compared to CNNs \cite{parkvision}. If the baseline Swin Transformer model is already finding a relatively flat minimum, the problem of getting stuck in sharp, suboptimal basins is less severe. Hence, the potential margin for improvement by VolSched is reduced.

\section{Discussion}
In the subsequent analysis, we examine the effect of VolSched by investigating the loss curves. Afterwards, we analyze the loss landscape of ResNet-18, with and without VolSched.

\subsection{Loss curve analysis}
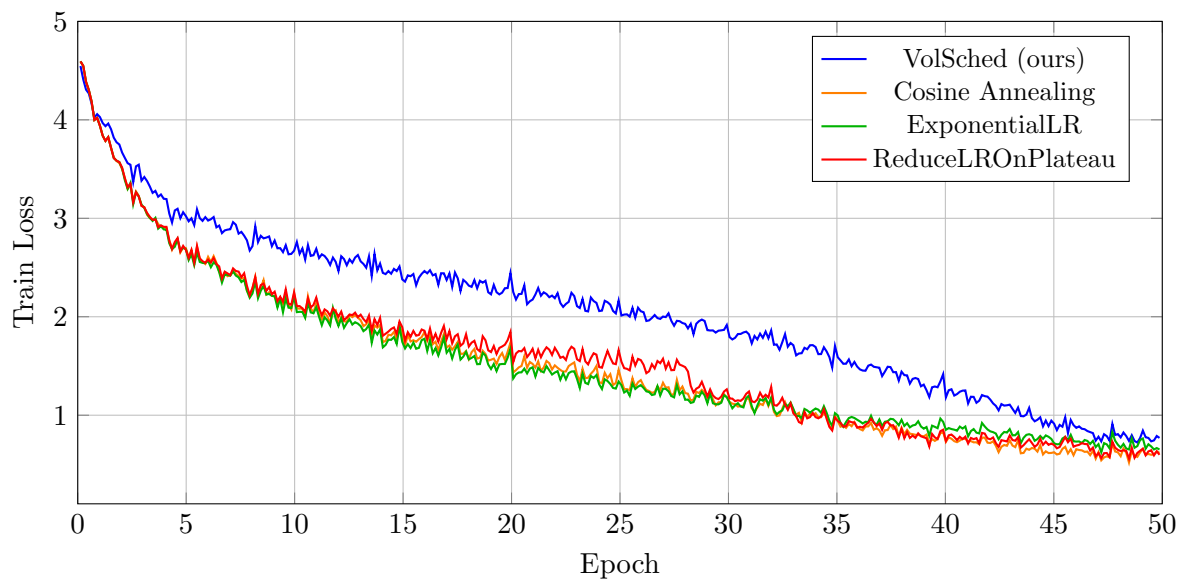
\begin{figure}[h!]
\centering
\begin{tikzpicture}
    \label{fig:losscurves}
    \begin{axis}[
        % --- General Axis Settings ---
        width=16cm,
        height=8cm,
        grid=major,
        %
        % --- Axis Labels and Title ---
        xlabel={Epoch},
        ylabel={Train Loss},
        %
        % --- Ticks and Limits ---
        xmin=0, xmax=50,
        ymin=0.1, ymax=5,
        %
        % --- Legend ---
        legend pos=north east,
        legend style={font=\small},
    ]

    % --- Plotting the Data from the CSV file ---

    % 1. Cosine Annealing (Blue)
    \addplot[no marks, blue, thick] 
    table [x=step, y=my_scheduler, col sep=comma] {losses.csv};
    \addlegendentry{VolSched (ours)}

    % 2. My Scheduler (Orange)
    \addplot[no marks, orange, thick] 
    table [x=step, y=cosine, col sep=comma] {losses.csv};
    \addlegendentry{Cosine Annealing}

    % 3. Exponential Scheduler (Green)
    \addplot[no marks, green!70!black, thick] 
    table [x=step, y=exponential, col sep=comma] {losses.csv};
    \addlegendentry{ExponentialLR}

    % 4. Plateau Scheduler (Red)
    \addplot[no marks, red, thick] 
    table [x=step, y=plateau, col sep=comma] {losses.csv};
    \addlegendentry{ReduceLROnPlateau}
    \end{axis}
\end{tikzpicture}
\caption{The plot shows the train loss on CIFAR-100 with various LR schedulers}
\end{figure}
In order to better understand how VolSched allows models to obtain superior test accuracy, we plot the train loss curves of ResNet-18 trained on CIFAR-100 for 50 epochs. A striking difference in the loss curves of VolSched and all other schedulers is observed in \hyperref[fig:losscurves]{Figure 3}. ReduceLROnPlateau, ExponentialLR, and Cosine Annealing all exhibit a rapid decrease in training loss during the initial epochs, while VolSched maintains a significantly higher loss. 

Cosine Annealing, ReduceLROnPlateau, and ExponentialLR, all display a common pattern of convergence. By epoch 35, the improvements are minimal. The low-drift nature of the loss curves after this point suggests that the models are no longer making substantial progress and are instead stuck within a local minima.

Meanwhile, the VolSched curve remains relatively high, hovering around a comparatively higher loss for most of the training, before dropping later on. The initial phase of high training loss can be interpreted as exploration, where a higher learning rate prevents the model from prematurely converging to a sharp or suboptimal local minimum. This resembles the behaviour of Sharpness Aware Minimization (SAM) \cite{sam}. However, while SAM explicitly optimizes for flat minima by solving a min–max problem, VolSched achieves this implicitly by using a dynamic learning rate to escape sharp valleys and explore more broadly, mimicking the effect of SAM's optimization methods.

The drop in loss observed in later epochs can be interpreted as a shift from exploration to exploitation. We hypothesize that by intentionally delaying this convergence, VolSched enables the model to discover a wider and flatter minimum, which is often correlated with better generalization performance to test data.
\iffalse
\subsection{Loss surface analysis}
\begin{figure}[ht]
    \label{fig:losslandscape}
    \centering
    \begin{minipage}[t]{0.48\linewidth}
        \centering
        VolSched
        \vspace{0.5em}
        \includegraphics[width=\linewidth]{image.png}
    \end{minipage}%
    \hfill % Pushes the two minipages apart to fill the line
    \begin{minipage}[t]{0.48\linewidth}
        \centering
        Cosine Annealing
        \vspace{0.5em}
        \includegraphics[width=\linewidth]{image2.png}
    \end{minipage}

    \caption{Training loss landscapes visualized around the trained ResNet-18 models. VolSched enables the model to find a smoother minima, leading to improved generalization performance.}
\end{figure}
To understand how VolSched obtains a superior local minima, we visualized the training loss landscape of ResNet-18 trained on CIFAR-100 for 50 epochs. Following the procedure in \cite{losslandscape}, we sampled two random directions, and evaluated the loss on a $20\times20$ grid (\hyperref[fig:losslandscape]{Figure 4}).

While the absolute minimum value of the loss function is comparable or even slightly deeper for the baseline, the landscape of the VolSched trained model exhibits a very flat basin at the center, with loss remaining quite low even with larger perturbations. Meanwhile, the Cosine Annealing landscape is markedly sharper, with the loss increasing rapidly when moving away from the center. The smoother surface implies a wider and flatter minimum as previously hypothesized, resulting in superior test performance.
\fi

\subsection{Loss landscape analysis}
To understand how VolSched achieves superior generalization, we move beyond the loss curves and analyze the geometry of the final model found by each scheduler trained with ResNet-18 for 50 epochs.

It is a widely held hypothesis that the flatness of a loss minimum is correlated with generalization performance, that flatter minima tend to generalize better than sharper ones \cite{flatminima}. A sharp minimum implies that small perturbations to the model's weights can cause a large increase in the loss, suggesting the model has overfit to the training data. Conversely, a flat minimum is more robust to such perturbations.

The curvature of the loss landscape at a given point is mathematically described by the Hessian matrix. The eigenvalues of the Hessian represent the curvature along its principal directions. A large top eigenvalue ($\lambda_{\max}$) indicates a sharp minimum, as there exists at least one direction in the parameter space where the loss increases rapidly.

Calculating the full Hessian is computationally infeasible for modern neural networks. Instead, we use the power iteration method to efficiently and accurately estimate the top eigenvalue of the Hessian at the final converged weights for each model \cite{eigenthings}. This provides a direct, quantitative measure of the sharpness of the solution.

\begin{table}[htbp]
\centering
\begin{tabular}{@{}lcccc@{}}
\toprule
& \thead{VolSched} & \thead{Cosine\\Annealing} & \thead{ExponentialLR} & \thead{ReduceLROnPlateau} \\
\midrule
$\lambda_{\max}$ & \textcolor{magenta}{\textbf{2.14$\pm$0.2}} & 3.94$\pm$0.3 & 5.12$\pm$0.7 & 3.45$\pm$0.5 \\
Accuracy & \textcolor{magenta}{\textbf{70.89$\pm$0.3}} & 69.31$\pm$0.2 & 67.35$\pm$0.5 & 69.47$\pm$0.7 \\
\bottomrule
\end{tabular}
\caption{Largest eigenvalues of Hessian matrices of ResNet-18 trained with various schedulers. \textcolor{magenta}{\textbf{Smallest eigenvalue}} and \textcolor{magenta}{\textbf{best accuracy}} are highlighted.}
\label{tab:results}
\end{table}

The VolSched-trained model converged to a minimum with a top eigenvalue of 2.14$\pm$0.2, which is 38\% flatter than the next-flattest minimum, found by ReduceLROnPlateau ($\lambda_{\max} = 3.45\pm0.5$). The other schedulers, ExponentialLR and Cosine Annealing, found even sharper minima, consistent with their lower test accuracy.

This result strongly supports our central claim. The prolonged exploration phase induced by VolSched, as observed in the loss curves \hyperref[fig:losscurves]{(Figure 3)}, prevents the model from prematurely converging to a sharp, overfitted local minimum. By maintaining a higher learning rate for longer, it enables the optimizer to discover a wider and significantly flatter basin in the loss landscape. This flatter minimum is more robust, leading directly to the observed improvements in generalization performance, providing concrete evidence for the mechanism behind VolSched's success.

\section{Conclusion}
In this paper, we proposed VolSched, a volatility-based adaptive learning rate scheduler that dynamically adjusts the learning rate based on past training histories, allowing the model to find a flatter minima, achieving improved generalization. Our experiments on the CIFAR-100 dataset show that VolSched consistently creates a more effective learning rate schedule compared to conventional methods. This resulted in a statistically significant gain of up to 1.4 percentage points on a competitive benchmark, achieved simply by replacing the scheduler. 

Our quantitative analysis of the loss landscape curvature further supports this conclusion. By measuring the top eigenvalue of the Hessian, we demonstrated that VolSched consistently finds significantly flatter minima than conventional methods. This suggests the performance gain stems from a more prolonged exploration phase which enables the discovery of wider, more robust, and more generalizable solutions.

\subsection{Limitations and future work}
While this study provides strong evidence for VolSched's efficacy, there are clear avenues for future research. Our experiments were performed on image classification. Future work should investigate the broader applicability of VolSched to different domains, like Natural Language Processing.

Furthermore, scalability experiments should also be conducted, with larger models being trained on larger datasets like ImageNet \cite{imagenet}. Such experiments would help to prove the efficacy of VolSched and its applicability in real world scenarios.

\newpage

\bibliography{citations}

\begin{thebibliography}{12}
\providecommand{\natexlab}[1]{#1}
\providecommand{\url}[1]{\texttt{#1}}
\expandafter\ifx\csname urlstyle\endcsname\relax
  \providecommand{\doi}[1]{doi: #1}\else
  \providecommand{\doi}{doi: \begingroup \urlstyle{rm}\Url}\fi

\bibitem[rec()]{recipes}
vision/references/classification at main · pytorch/vision --- github.com.
\newblock \url{https://github.com/pytorch/vision/tree/main/references/classification#resnet}.

\bibitem[red()]{reducelr}
{R}educe{L}{R}{O}n{P}lateau; {P}y{T}orch 2.7 documentation --- docs.pytorch.org.
\newblock \url{https://docs.pytorch.org/docs/stable/generated/torch.optim.lr_scheduler.ReduceLROnPlateau.html}.

\bibitem[Cubuk et~al.(2018)Cubuk, Zoph, Mane, Vasudevan, and Le]{autoaug}
Ekin~D Cubuk, Barret Zoph, Dandelion Mane, Vijay Vasudevan, and Quoc~V Le.
\newblock Autoaugment: Learning augmentation policies from data.
\newblock \emph{arXiv preprint arXiv:1805.09501}, 2018.

\bibitem[Deng et~al.(2009)Deng, Dong, Socher, Li, Li, and Fei-Fei]{imagenet}
Jia Deng, Wei Dong, Richard Socher, Li-Jia Li, Kai Li, and Li~Fei-Fei.
\newblock Imagenet: A large-scale hierarchical image database.
\newblock In \emph{2009 IEEE Conference on Computer Vision and Pattern Recognition}, pages 248--255, 2009.
\newblock \doi{10.1109/CVPR.2009.5206848}.

\bibitem[Foret et~al.(2021)Foret, Kleiner, Mobahi, and Neyshabur]{sam}
Pierre Foret, Ariel Kleiner, Hossein Mobahi, and Behnam Neyshabur.
\newblock Sharpness-aware minimization for efficiently improving generalization, 2021.
\newblock URL \url{https://arxiv.org/abs/2010.01412}.

\bibitem[Golmant et~al.(2018)Golmant, Yao, Gholami, Mahoney, and Gonzalez]{eigenthings}
Noah Golmant, Zhewei Yao, Amir Gholami, Michael Mahoney, and Joseph Gonzalez.
\newblock pytorch-hessian-eigenthings: efficient pytorch hessian eigendecomposition, October 2018.
\newblock URL \url{https://github.com/noahgolmant/pytorch-hessian-eigenthings}.

\bibitem[He et~al.(2016)He, Zhang, Ren, and Sun]{resnet}
Kaiming He, Xiangyu Zhang, Shaoqing Ren, and Jian Sun.
\newblock Deep residual learning for image recognition.
\newblock In \emph{Proceedings of the IEEE conference on computer vision and pattern recognition}, pages 770--778, 2016.

\bibitem[Keskar et~al.(2016)Keskar, Mudigere, Nocedal, Smelyanskiy, and Tang]{flatminima}
Nitish~Shirish Keskar, Dheevatsa Mudigere, Jorge Nocedal, Mikhail Smelyanskiy, and Ping Tak~Peter Tang.
\newblock On large-batch training for deep learning: Generalization gap and sharp minima.
\newblock \emph{arXiv preprint arXiv:1609.04836}, 2016.

\bibitem[Liu et~al.(2021)Liu, Lin, Cao, Hu, Wei, Zhang, Lin, and Guo]{swin}
Ze~Liu, Yutong Lin, Yue Cao, Han Hu, Yixuan Wei, Zheng Zhang, Stephen Lin, and Baining Guo.
\newblock Swin transformer: Hierarchical vision transformer using shifted windows.
\newblock In \emph{2021 IEEE/CVF International Conference on Computer Vision (ICCV)}, pages 9992--10002. IEEE, 2021.

\bibitem[Loshchilov and Hutter(2017{\natexlab{a}})]{adamw}
Ilya Loshchilov and Frank Hutter.
\newblock Decoupled weight decay regularization.
\newblock \emph{arXiv preprint arXiv:1711.05101}, 2017{\natexlab{a}}.

\bibitem[Loshchilov and Hutter(2017{\natexlab{b}})]{cosineanneal}
Ilya Loshchilov and Frank Hutter.
\newblock Sgdr: Stochastic gradient descent with warm restarts, 2017{\natexlab{b}}.
\newblock URL \url{https://arxiv.org/abs/1608.03983}.

\bibitem[Park and Kim(2022)]{parkvision}
Namuk Park and Songkuk Kim.
\newblock How do vision transformers work?
\newblock In \emph{10th International Conference on Learning Representations, ICLR 2022}, 2022.

\end{thebibliography}
\newpage
\appendix
\section{Experimental setup for ResNet-18 and ResNet-34}
\label{sec:exp}
\begin{itemize}
    \item \textbf{Batch size:} 128\vspace{-5pt}
    \item \textbf{Data augmentation:} RandomCrop, RandomHorizontalFlip, AutoAugment\vspace{-5pt}
    \item \textbf{Number of workers:} 4\vspace{-5pt}
    \item \textbf{Optimizer:} SGD with initial learning rate 1e-1, momentum 0.9, weight decay 1e-4. Nesterov momentum disabled.\vspace{-5pt}
    \item \textbf{Repetitions:} 3 independent runs per configuration with seeds [8, 42, 123]. In the case of 50 epoch ResNet-18, 5 independent runs with seeds [8, 42, 123, 1234, 12345] were performed.\vspace{-5pt}
    \item \textbf{eta\_min:} 1e-4\vspace{-5pt}
    \item \textbf{T\_max:} epochs $\times$ steps per epoch\vspace{-5pt}
    \item \textbf{Hardware:} 1x NVIDIA Tesla P100-PCIE-16GB GPU (CUDA 12.6, Driver 560.35.03)\vspace{-5pt}
    \item \textbf{VolSched hyperparameters:} $w = 0.05$ and $w = 0.03$ for ResNet-18 and ResNet-34 respectively, $N = 50$\vspace{-5pt}
    \item \textbf{ExponentialLR hyperparameters:} $\gamma = 0.95$, warmup\_epochs = 1, start\_factor = 0.01\vspace{-5pt}
    \item \textbf{ReduceLROnPlateau hyperparameters:} mode = max, factor = 0.5, patience = 10, warmup\_epochs = 1, start\_factor = 0.01\vspace{-5pt}
    \item \textbf{Cosine Annealing hyperparameters:} warmup\_epochs = 1, start\_factor = 0.01\vspace{-5pt}
    \item \textbf{RandomCrop configuration:} size = 32, padding = 4\vspace{-5pt}
    \item \textbf{AutoAugment configuration:} Learned CIFAR-10 policy
\end{itemize}

\section{Experimental setup for Swin ViT}
\label{sec:exp2}
Only items differing from ResNet tests are stated.
\begin{itemize}
    \item \textbf{Batch size:} 1024\vspace{-5pt}
    \item \textbf{Optimizer:} AdamW with learning rate 1e-3, weight decay 1e-4, betas 0.9, 0.999\vspace{-5pt}
    \item \textbf{Repetitions:} 4 independent runs per configuration with seeds [8, 42, 123, 1234].\vspace{-5pt}
    \item \textbf{Epochs:} 80\vspace{-5pt}
    \item \textbf{eta\_min:} 1e-5\vspace{-5pt}
    \item \textbf{Hardware:} 2x NVIDIA Tesla T4 GPUs (CUDA 12.6, Driver 560.35.03)\vspace{-5pt}
    \item \textbf{VolSched hyperparameters:} $w = 0.03$, $N = 25$, warmup\_epochs = 1, start\_factor = 0.01\vspace{-5pt}
\end{itemize}

\section{Hyperparameter sensitivity analysis}
\label{sec:hsa}

Using the same configuration, we conducted hyperparameter sensitivity analyses for VolSched with ResNet-18 on CIFAR-100. We vary $w$, setting it to 0.01, 0.05, and 0.1.

\begin{table}[htbp]
\centering
\begin{tabular}{lccccc}
\toprule
\thead{Models} & \thead{Epochs} & \thead{Baseline} & \thead{$w = 0.01$} & \thead{$w = 0.05$} & \thead{$w = 0.1$} \\
\midrule
ResNet-18 & 50 & 69.37$\pm$0.2 & 69.39$\pm$0.3 & \textcolor{magenta}{\textbf{70.76$\pm$0.4}} & 62.6$\pm$7.6 \\
\bottomrule
\end{tabular}
\caption{Top-1 accuracy (\%) of ResNet-18 models on CIFAR-100, with varying $w$. \textcolor{magenta}{\textbf{Best model}} is highlighted.}
\label{tab:hsa}
\end{table}

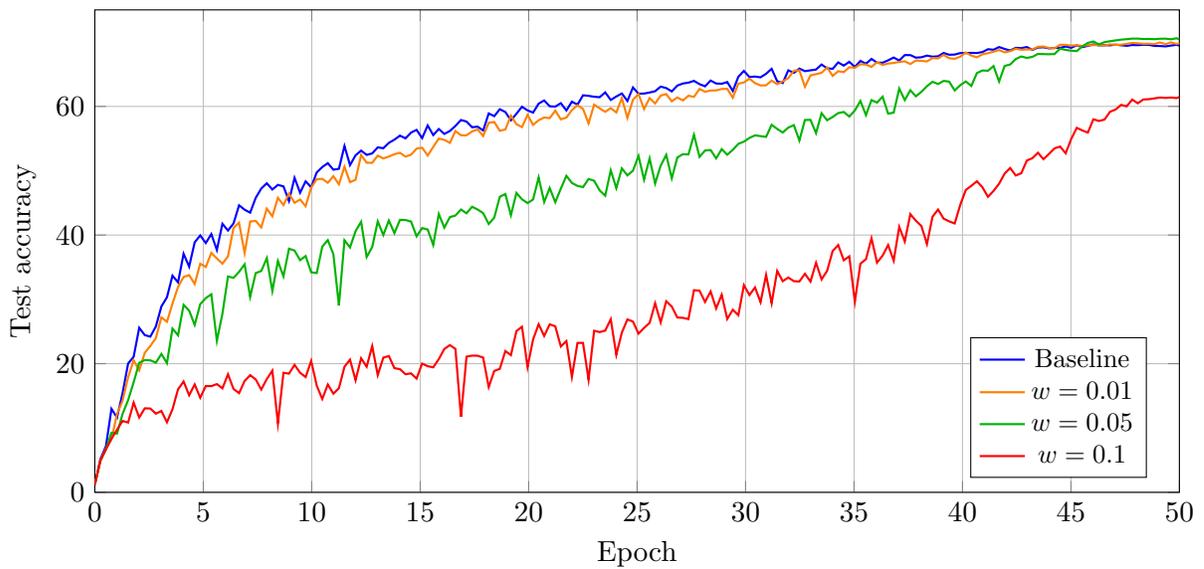
\begin{figure}[h!]
\label{fig:hsa}
\centering
\begin{tikzpicture}
    \begin{axis}[
        % --- General Axis Settings ---
        width=16cm,
        height=8cm,
        grid=major,
        %
        % --- Axis Labels and Title ---
        xlabel={Epoch},
        ylabel={Test accuracy},
        %
        % --- Ticks and Limits ---
        xmin=0, xmax=50,
        ymin=0, ymax=75,
        %
        % --- Legend ---
        legend pos=south east,
        legend style={font=\small},
    ]

    % --- Plotting the Data from the CSV file ---

    % 1. Cosine Annealing (Blue)
    \addplot[no marks, blue, thick] 
    table [x=step, y=cosine, col sep=comma] {data2.csv};
    \addlegendentry{Baseline}

    % 2. My Scheduler (Orange)
    \addplot[no marks, orange, thick] 
    table [x=step, y=exponential, col sep=comma] {data2.csv};
    \addlegendentry{$w = 0.01$}

    % 3. Exponential Scheduler (Green)
    \addplot[no marks, green!70!black, thick] 
    table [x=step, y=my_scheduler, col sep=comma] {data2.csv};
    \addlegendentry{$w = 0.05$}

    % 4. Plateau Scheduler (Red)
    \addplot[no marks, red, thick] 
    table [x=step, y=plateau, col sep=comma] {data2.csv};
    \addlegendentry{$w = 0.1$}

    \end{axis}
\end{tikzpicture}
\caption{The plot shows the test accuracy on CIFAR-100 with VolSched with varying $w$}
\end{figure}

\begin{figure}[h!]
\centering
\label{fig:hsalr}
\begin{tikzpicture}
    \begin{axis}[
        % --- General Axis Settings ---
        width=16cm,
        height=8cm,
        grid=major,
        %
        % --- Axis Labels and Title ---
        xlabel={Epoch},
        ylabel={Learning rate},
        %
        % --- Ticks and Limits ---
        xmin=0, xmax=50,
        ymin=0, ymax=1.7,
        %
        % --- Legend ---
        legend pos=north east,
        legend style={font=\small},
    ]

    % --- Plotting the Data from the CSV file ---

    % 1. Cosine Annealing (Blue)
    \addplot[no marks, blue, thick] 
    table [x=step, y=cosine, col sep=comma] {data3.csv};
    \addlegendentry{Baseline}

    % 2. My Scheduler (Orange)
    \addplot[no marks, orange, thick] 
    table [x=step, y=exponential, col sep=comma] {data3.csv};
    \addlegendentry{$w = 0.01$}

    % 3. Exponential Scheduler (Green)
    \addplot[no marks, green!70!black, thick] 
    table [x=step, y=my_scheduler, col sep=comma] {data3.csv};
    \addlegendentry{$w = 0.05$}

    % 4. Plateau Scheduler (Red)
    \addplot[no marks, red, thick] 
    table [x=step, y=plateau, col sep=comma] {data3.csv};
    \addlegendentry{$w = 0.1$}

    \end{axis}
\end{tikzpicture}
\caption{The plot shows the learning rates of the model with VolSched with varying $w$}
\end{figure}
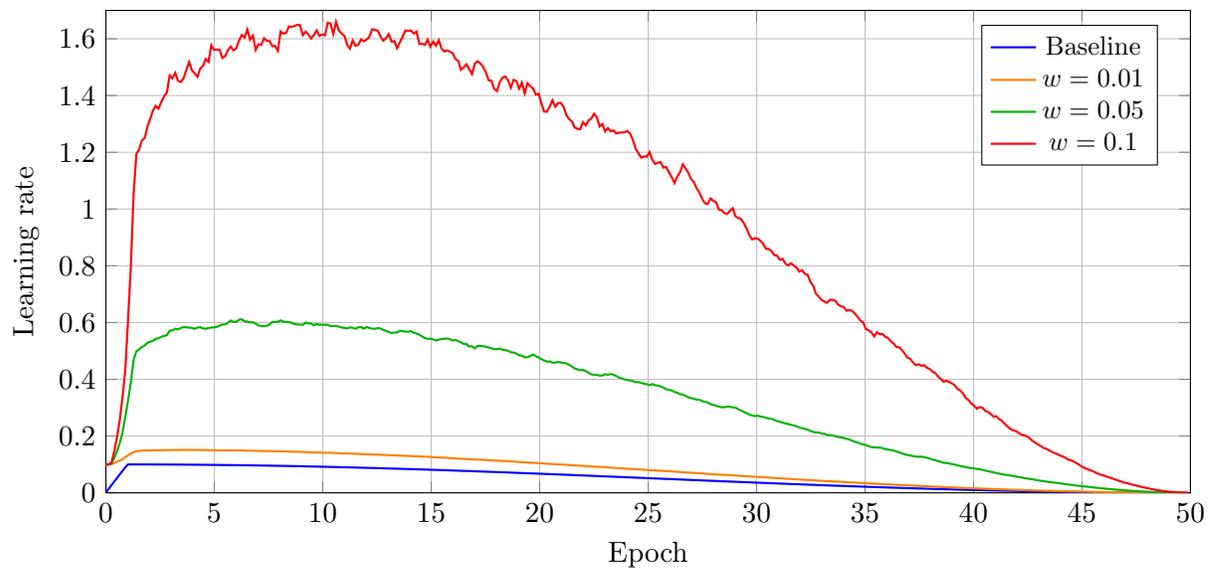
As can be seen in Figures \hyperref[fig:hsa]{4} and \hyperref[fig:hsalr]{5}, setting $w$ to a value too low will often lead to minimal benefits above the baseline of Cosine Annealing, while setting $w$ too high will lead to overly high learning rates, resulting in unstable training, and potentially model divergence.

\end{document}